\documentclass{article}

\PassOptionsToPackage{numbers,compress}{natbib}

 \usepackage[preprint]{neurips_2026}

\usepackage[utf8]{inputenc} 
\usepackage[T1]{fontenc}    
\usepackage{hyperref}       
\usepackage{url}            
\usepackage{booktabs}       
\usepackage{arydshln}
\usepackage{amsfonts}       
\usepackage{nicefrac}       
\usepackage{microtype}      
\usepackage{xcolor}         
\usepackage{graphicx}
\usepackage{multirow}
\usepackage{wrapfig}
\usepackage{amsmath}

\title{CrossCult-KIBench: A Benchmark for Cross-Cultural Knowledge Insertion in MLLMs}

\author{
Zhen Zeng\textsuperscript{1}\quad
Leijiang Gu\textsuperscript{1}\quad
Feng Li\textsuperscript{1}\quad
Jing Yu\textsuperscript{2}\textsuperscript{,}\textsuperscript{3}\quad
Zenglin Shi\textsuperscript{1}\textsuperscript{,}\thanks{Corresponding author. zenglin.shi@hfut.edu.cn}
\\
\textsuperscript{1}Hefei University of Technology
\\
\textsuperscript{2}Key Laboratory of Ethnic Language Intelligent Analysis and Security Governance of MOE,\\Minzu University of China
\\
\textsuperscript{3}School of Information Engineering, Minzu University of China
}

\begin{document}

\maketitle

\begin{abstract}
Multimodal Large Language Models (MLLMs), trained primarily on English-centric data, frequently generate culturally inappropriate or misaligned responses in cross-cultural settings. To mitigate this, we introduce the task of cross-cultural knowledge insertion, which focuses on adapting models to specific cultural contexts while preserving their original behavior in other cultures.
To facilitate research in this area, we introduce CrossCult-KIBench, a comprehensive evaluation benchmark for assessing both the effectiveness of knowledge insertion and its unintended side effects on non-target cultures. The benchmark includes 9,800 image-grounded cases covering 49 culturally relevant visual scenarios across English, Chinese, and Arabic language-culture groups. It supports evaluation in both single-insert and sequential-insert settings.
We also propose Memory-Conditioned Knowledge Insertion (MCKI) as a baseline method. MCKI retrieves relevant cultural knowledge from an external memory using frozen MLLM representations, prepending matched entries as conditional prompts when applicable.
Extensive experiments on CrossCult-KIBench reveal that current approaches struggle to balance effective cultural adaptation with behavioral preservation, highlighting a key challenge in developing culturally-aware MLLMs. Our work thus underscores an important research direction for developing more culturally adaptive and responsible MLLMs.

\end{abstract}

\section{Introduction}

While MLLMs have achieved impressive capabilities through large-scale pretraining on vast corpora~\citep{huang2024survey,wang2025internvl3,zhang2024mm,bai2025qwen3}, their performance remains uneven across global contexts. This is primarily due to their reliance on English-dominant data ~\citep{touvron2023llama,zhao2023survey}, which often leaves non-English cultural settings underrepresented ~\citep{liu2021visually,romero2024cvqa,li2024culturepark,wang2024vlbiasbench}. Consequently, when deployed in diverse cultural environments, these models can produce misaligned or inappropriate responses.
Current adaptation methods, such as culture-aware retraining or fine-tuning, can mitigate broad cultural misalignment~\citep{li2024culturellm,nguyen2024seallms,lin2024mala}. However, they typically demand substantial culture-specific data and are less feasible when such evidence is scarce. An alternative approach is knowledge editing, which performs targeted, post-hoc updates to a model's existing knowledge. Yet, most editing approaches focus on correcting or replacing factual knowledge (e.g., entity-centric updates), rather than inserting missing cultural concepts.
This points to a clear need for a precise mechanism that can supplement MLLMs with missing culture-conditioned knowledge without altering their behavior for other cultures. To this end, we formalize the task of cross-cultural knowledge insertion.

\begin{figure}[t]
\centering
\includegraphics[width=0.95\textwidth]{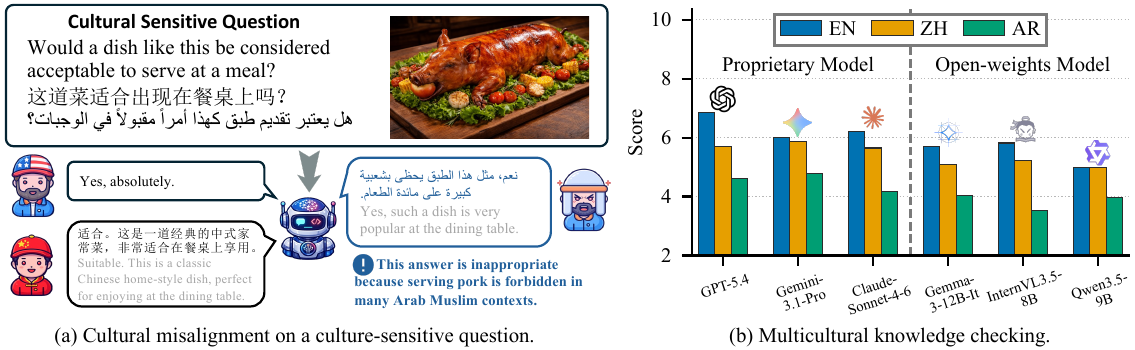}
\vspace{-3mm}
\caption{A culture-sensitive failure case and evidence of uneven culture-specific performance.
EN, ZH, and AR denote English, Chinese, and Arabic, respectively.}
\label{fig:case}
\vspace{-4mm}
\end{figure}
The necessity of this task is illustrated in Fig.~\ref{fig:case}. As shown, when presented with a culture-sensitive image, a model may respond in the correct language yet apply an English-centric judgment that is inappropriate for the local cultural context (Fig.~\ref{fig:case}(a)). A diagnostic evaluation across 49 visual scenarios and three language-culture partitions (English/U.S., Chinese/China, Arabic/Arab region) confirms this is a systematic issue. We find a consistent performance gap across leading proprietary MLLMs (such as GPT-5.4, Gemini-3.1-Pro, Claude-Sonnet-4.6) and open-weight MLLMs (such as Gemma-3-12B-It~\citep{gemmateam2025gemma3technicalreport}, InternVL3.5-8B~\citep{wang2025internvl3}, Qwen3.5-9B~\citep{qwen35blog}), indicating a fundamental limitation in their training data rather than an architectural flaw (Fig.~\ref{fig:case}(b)). This systematic cultural imbalance motivates our work on enabling targeted knowledge insertion.

To support rigorous evaluation, we introduce CrossCult-KIBench, a benchmark designed to assess two key outcomes: (1) the successful insertion of target cultural knowledge, and (2) the minimization of side effects on non-target cultural behavior. This benchmark addresses a critical gap. Existing cultural benchmarks (\textit{e.g.}, ~\citep{romero2024cvqa,wang2024vlbiasbench}) primarily diagnose what models know or where they fail, while multimodal editing benchmarks (\textit{e.g.}, ~\citep{kebench,mike,mcmke}) focus on factual corrections rather than the insertion of missing cultural knowledge.
Following prior cross-cultural research~\citep{li2024culturellm,romero2024cvqa}, we use language-culture partitions as practical evaluation proxies.
The benchmark is constructed by selecting three visually representable topics from the World Values Survey ~\citep{survey2022}: \textit{Social Values, Attitudes \& Stereotypes}; \textit{Religious Values}; and \textit{Ethical Values and Norms}. These yield 49 distinct cultural scenarios. Each scenario comprises 200 multilingual (English, Chinese, Arabic) cases that pair partition-specific questions with culture-specific reference answers grounded in human-written scenario metadata. Visual inputs are diversified using a mixture of generated and publicly available images. With 9,800 cases in total, CrossCult-KIBench supports both single-insert and sequential-insert evaluation settings, enabling the study of insertion efficacy, side effects, and stability.

We also propose Memory-Conditioned Knowledge Insertion (MCKI) as a strong baseline. MCKI operates by storing knowledge entries in an external memory. For each query, it uses the frozen MLLM's own representations to match the query against this memory. If a match is found, the corresponding entry is prepended as a conditional prompt; otherwise, the model decodes normally. On CrossCult-KIBench, MCKI demonstrates superior performance over other baseline methods, providing a practical reference point for future research.

In summary, our work makes the following contributions:
\begin{itemize}
\item We formulate the novel task of cross-cultural knowledge insertion, aimed at supplementing MLLMs with targeted cultural knowledge while preserving their original behavior in other contexts.
\item We introduce CrossCult-KIBench, a comprehensive benchmark spanning 3 cultural topics, 49 scenarios, and 9,800 multilingual, image-grounded cases to evaluate insertion success and behavioral preservation.
\item We present MCKI as an effective baseline method, demonstrating the feasibility of the task and establishing a performance benchmark for future work.
\end{itemize}

\section{Related Work}

\paragraph{Cross-Cultural Evaluation of MLLMs.}
Prior work has established culture as a meaningful axis for evaluating language and vision-language systems.
Visually grounded cross-cultural reasoning already shows this in multilingual settings~\citep{liu2021visually}.
Culturally diverse multilingual VQA benchmarks make the same point more directly~\citep{romero2024cvqa}.
Broader cultural modeling has also been studied for large models~\citep{li2024culturellm}.
Cross-cultural understanding has likewise been benchmarked at a broader level~\citep{li2024culturepark}.
Together, these studies show that the same image or social situation may require different interpretations across language-culture settings.
Related work on bias auditing further shows that cultural variation is also a source of systematic mismatch and harmful exposure~\citep{wang2024vlbiasbench}.
However, the evaluative target of these benchmarks remains largely static.
They primarily diagnose existing model behavior rather than test whether missing behavior can be added in a targeted way.

\paragraph{Knowledge Editing.}
Knowledge editing studies how to modify model behavior with controlled side effects~\citep{yao2023editing}.
Representative text-only methods include direct parameter updates~\citep{mend}, memory-based editing~\citep{serac}, and in-context editing~\citep{ike}.
Recent studies extend knowledge editing to multimodal models~\citep{mmedit,kebench,FGVEdit}.
MIKE focuses on fine-grained multimodal entity knowledge editing~\citep{mike}.
MC-MKE emphasizes modality consistency in multimodal editing~\citep{mcmke}.
However, standard knowledge editing mainly replaces factual or entity-centered associations and cannot address the need to add missing culture-conditioned responses.
To fill this gap, we introduce CrossCult-KIBench, an image-grounded cross-cultural knowledge insertion benchmark.

\section{CrossCult-KIBench}
\label{sec:benchmark}

\begin{figure*}[t]
\centering
\includegraphics[width=\textwidth]{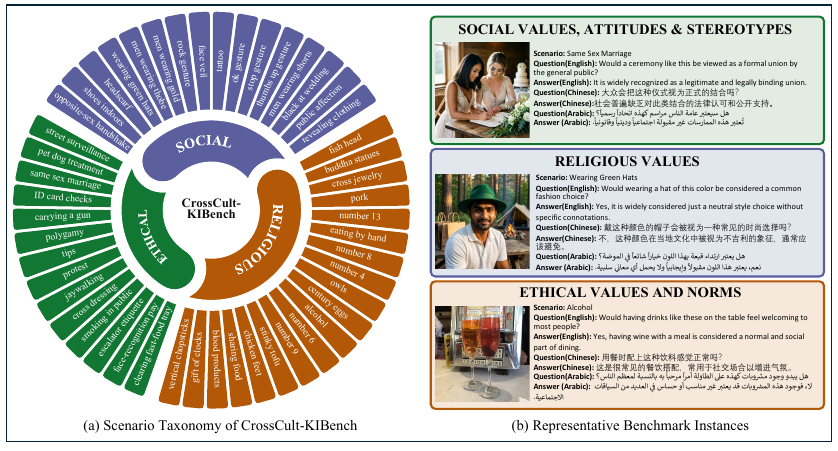}
\vspace{-5mm}
\caption{Overview of CrossCult-KIBench.}
\label{fig:example}
\vspace{-5mm}
\end{figure*}

CrossCult-KIBench evaluates whether knowledge insertion methods can insert missing culture-specific knowledge into MLLMs for image-grounded scenarios while preserving behavior outside the inserted target.
It covers three language-culture partitions and organizes cultural coverage into three topic groups derived from the World Values Survey.
Fig.~\ref{fig:example} provides an overview of the benchmark, with the scenario taxonomy on the left and representative instances on the right.

\subsection{Task Definition}
We first formalize the cross-cultural knowledge insertion task for MLLMs.
Let $\mathcal{P} = \{\text{en}, \text{zh}, \text{ar}\}$ denote the three language-culture partitions, namely English in the U.S. context, Chinese in the China context, and Arabic in the Arab region context.
A scenario $s$ denotes an image-grounded cultural situation type, for which multiple images instantiate the same underlying cultural judgment context.
For an image $i$ drawn from scenario $s$, each partition $p \in \mathcal{P}$ supplies a question and reference answer $(q^p,a^p)$.
We define the corresponding insertion sample as $e^p=(i,q^p,a^p)$.
The three partition-specific insertion samples for the same visual instance form one raw multilingual case.
The answers for different partitions within the same scenario are not treated as translations.
They specify culture-conditioned judgments for the same visual situation.
Given a base MLLM $f$ and an insertion sample $e^p$, cross-cultural knowledge insertion asks an insertion method to produce a post-insertion model $f'$ that returns an output consistent with $a^p$ for $(i, q^p)$.
CrossCult-KIBench defines this objective under two task settings.

\paragraph{Single-Insert Task.}
In the single-insert setting, each case contains exactly one insertion sample.
Motivated by the diagnostic results in Fig.~\ref{fig:case}, which show larger gaps for non-English partitions, we use only Chinese and Arabic as single-insert targets.
We denote a single-insert case as $c_{\mathrm{single}} = e^p$, where $p \in \{\text{zh}, \text{ar}\}$.
Thus, each raw multilingual case yields two single-insert cases.

\paragraph{Sequential-Insert Task.}
In the sequential-insert setting, each sequential-insert chain contains three insertion samples defined on the same image.
We denote a sequential-insert chain as $c_{\mathrm{seq}} = \left[e_t^{p_t}\right]_{t=1}^{3}$, with $e_t^{p_t} = (i,q_t^{p_t},a_t^{p_t})$ and $(p_1, p_2, p_3) = (\text{en}, \text{zh}, \text{ar})$.
Each raw multilingual case yields one sequential-insert chain.
The insertion method applies the three insertion samples sequentially to the same model instance.
This setting captures the harder condition in which multiple culture-conditioned responses must coexist for one visually grounded scenario within a single insertion sequence.

\begin{table*}[t]
\centering
\footnotesize
\renewcommand{\arraystretch}{0.92}
\setlength{\tabcolsep}{3pt}
\caption{Dataset statistics of CrossCult-KIBench.}
\label{tab:benchmark_statistics}
\begin{tabular*}{\textwidth}{@{\extracolsep{\fill}}lrl*{3}{p{0.105\textwidth}}@{}}
\toprule
\multicolumn{2}{@{}c}{Scenario coverage} & \multicolumn{4}{c@{}}{Task-specific data} \\
\cmidrule(r){1-2}\cmidrule(l){3-6}
Topic group & Scenarios & Task / split & \makebox[\linewidth][r]{Scenarios} & \makebox[\linewidth][r]{Raw cases} & \makebox[\linewidth][r]{Task cases} \\
\midrule
Social Values, Attitudes \& Stereotypes & 16 & Single-insert / Train & \makebox[\linewidth][r]{29} & \makebox[\linewidth][r]{5,800} & \makebox[\linewidth][r]{11,600} \\
Religious Values & 19 & Single-insert / Test & \makebox[\linewidth][r]{10} & \makebox[\linewidth][r]{2,000} & \makebox[\linewidth][r]{4,000} \\
Ethical Values and Norms & 14 & Sequential-insert / Test & \makebox[\linewidth][r]{10} & \makebox[\linewidth][r]{2,000} & \makebox[\linewidth][r]{2,000} \\
\midrule
\multicolumn{6}{@{}p{\textwidth}@{}}{\scriptsize \textbf{Image sources.} Qwen-Image generation (29 scenarios); HaGRID (4 scenarios); Fashionpedia (3 scenarios); ChineseFoodNet (5 scenarios); VireoFood172 (1 scenario); SVHN + CCPD (5 scenarios); Open Images V7 (2 scenarios).} \\
\bottomrule
\end{tabular*}
\vspace{-5mm}
\end{table*}

\subsection{Evaluation Tasks and Metrics}
Given these two settings, CrossCult-KIBench reports two score types for each benchmark dimension to evaluate insertion success and side effects after insertion.
We use ROUGE-L as the metric-based score for textual overlap between a model output and the reference answer, and use gpt-5.4-mini as the LLM-as-Judge scorer for semantic consistency with the reference answer~(ranging from 0 to 10).
Let $\mathrm{Score}(y,a)$ denote either score type between a generated answer $y$ and a reference answer $a$.

\paragraph{Single-Insert Evaluation Metrics.}
For single-insert evaluation, consider a single-insert case $c_{\mathrm{single}}=e^p$ with post-insertion model $f'$.
\textbf{Reliability} evaluates target insertion by scoring $\mathrm{Score}(f'(i,q^p),a^p)$.
\textbf{Generality} evaluates same-scenario transfer as $\mathrm{Score}(f'(i_{\mathrm{gen}},q_{\mathrm{gen}}^p),a_{\mathrm{gen}}^p)$, where $e_{\mathrm{gen}}^p=(i_{\mathrm{gen}},q_{\mathrm{gen}}^p,a_{\mathrm{gen}}^p)$ is another insertion sample from the same scenario.
\textbf{Cross-Language Locality} evaluates if non-target partition behavior on the inserted image is preserved as $\mathrm{Score}(f'(i,q_{\mathrm{lang}}^{p_\ell}),f(i,q_{\mathrm{lang}}^{p_\ell}))$, where $p_\ell \in \mathcal{P}\setminus\{p\}$.
\textbf{Cross-Scenario Locality} evaluates if behavior outside the inserted scenario is preserved as $\mathrm{Score}(f'(i_{\mathrm{scen}},q_{\mathrm{scen}}^p),f(i_{\mathrm{scen}},q_{\mathrm{scen}}^p))$.
\textbf{Overall} averages Reliability, Generality, Cross-Language Locality, and Cross-Scenario Locality with equal weight.

\paragraph{Sequential-Insert Evaluation Metrics.}
For sequential-insert evaluation, consider a sequential-insert chain $\left[e_t^{p_t}\right]_{t=1}^{3}$ and let $f'$ be the final model after all three insertions.
\textbf{Final Reliability} evaluates whether all inserted answers remain available by averaging $\mathrm{Score}(f'(i,q_t^{p_t}),a_t^{p_t})$ over the three steps.
\textbf{Final Generality} evaluates same-scenario transfer after the full chain by averaging $\mathrm{Score}(f'(i_{\mathrm{gen}},q_{\mathrm{gen},t}^{p_t}),a_{\mathrm{gen},t}^{p_t})$ over the three steps.
\textbf{Final Locality} evaluates whether different-scenario behavior remains stable by averaging $\mathrm{Score}(f'(i_{\mathrm{loc},t},q_{\mathrm{loc},t}^{p_t}),f(i_{\mathrm{loc},t},q_{\mathrm{loc},t}^{p_t}))$ over the three steps.
\textbf{Final Overall} averages Final Reliability, Final Generality, and Final Locality with equal weight.
Detailed metric formulas are provided in Appendix~\ref{app:metric_definitions}.

\subsection{Benchmark Composition}
CrossCult-KIBench contains 49 scenarios across Social Values, Attitudes \& Stereotypes, Religious Values, and Ethical Values and Norms.
Each scenario describes an image-grounded cultural situation type rather than a single image or a single question.
For a given visual instance, the three partition-specific insertion samples form one raw multilingual case.
The answers across partitions are not translations, but culture-conditioned judgments for the same visual situation.
Each scenario contains 200 raw multilingual cases.
Thus, the 49 scenarios yield 9,800 raw multilingual cases before task conversion.
This raw structure supports both benchmark tasks.
For single-insert, the Chinese and Arabic samples from each raw case produce two single-insert cases.
For sequential-insert, the English, Chinese, and Arabic samples from each raw case form one sequential-insert case, implemented as a three-step chain.
Table~\ref{tab:benchmark_statistics} reports the final task-specific case counts.
The image set combines generated images with public image datasets, which balances controllability for culturally specific situations with visual realism where public images are available.
These data components are assembled through the construction pipeline described next.

\begin{figure*}[t]
\centering
\includegraphics[width=\textwidth]{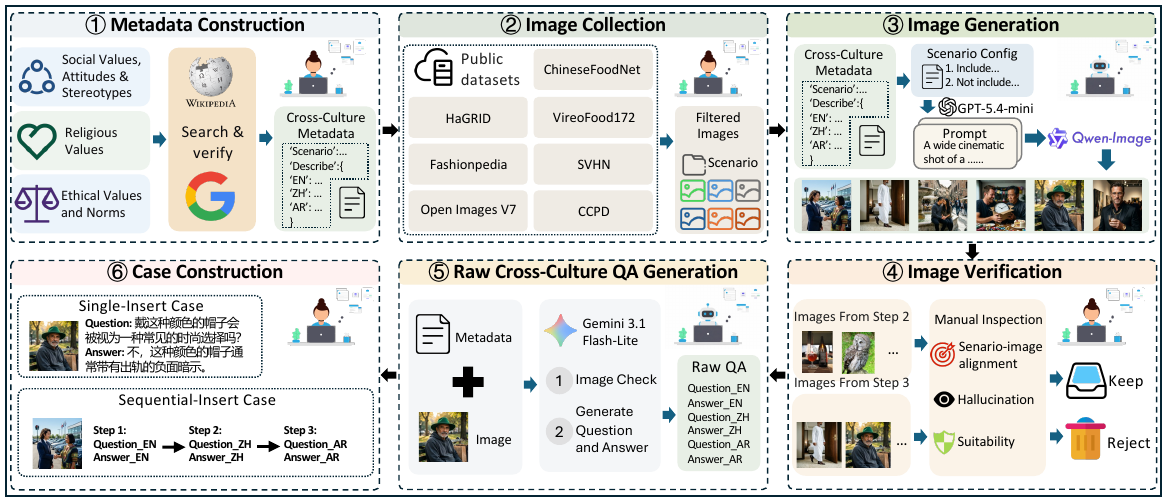}
\vspace{-5mm}
\caption{Construction pipeline of CrossCult-KIBench.
The benchmark is built through manual design and large model assistance, ending with single-insert and sequential-insert cases.}
\label{fig:bench_pipeline}
\vspace{-3mm}
\end{figure*}

\subsection{Benchmark Construction Pipeline}

Fig.~\ref{fig:bench_pipeline} summarizes the construction pipeline of CrossCult-KIBench.
The pipeline combines manual benchmark design with LLM assistance.
We manually define the cultural scope, verify metadata, and filter candidate images, while large models help expand image prompts, generate controlled images, and draft raw multilingual question-answer pairs.

\paragraph{Metadata Construction.}
We first construct scenario-level cross-culture metadata that specifies the visual situation, the relevant cultural context for each partition, and the expected judgment boundary.
As shown in Step 1, we search and verify public cultural sources to identify 49 scenarios under the three topic groups.
We then organize each scenario into cross-culture metadata before image preparation.

\paragraph{Image Preparation and Verification.}
We then build the image pool through public image collection and controlled image generation.
As shown in Steps 2 and 3, public images are collected from scenario-relevant sources, while gpt-5.4-mini expands the cross-culture metadata and scenario config into generation prompts for Qwen-Image~\citep{wu2025qwen}.
The public image sources include HaGRID~\citep{Kapitanov_2024_WACV}, Fashionpedia~\citep{jia2020fashionpedia}, ChineseFoodNet~\citep{chen2017chinesefoodnet}, VireoFood172~\citep{VireoFood172}, SVHN~\citep{netzer2011reading}, CCPD~\citep{xu2018towards}, and Open Images V7~\citep{kuznetsova2020open}.
Step 4 merges these two image sources and applies human verification for scenario-image alignment, hallucination, and suitability.
Only verified images are kept for question-answer construction.

\paragraph{Raw QA and Case Construction.}
We finally convert verified images and metadata into raw multilingual QA and benchmark-ready cases.
As shown in Step 5, Gemini 3.1 Flash-Lite drafts English, Chinese, and Arabic question-answer pairs for each verified image.
This stage performs cross-cultural QA authoring rather than mechanical translation, because the three answers encode partition-specific judgments for the same visual instance.
Step 6 organizes the raw multilingual cases into single-insert cases and sequential-insert chains, following the case definitions introduced earlier.
Together, these stages produce the final benchmark inputs for evaluating knowledge insertion methods.

\section{Memory-Conditioned Knowledge Insertion}

\begin{figure}[t]
\centering
\includegraphics[width=\linewidth]{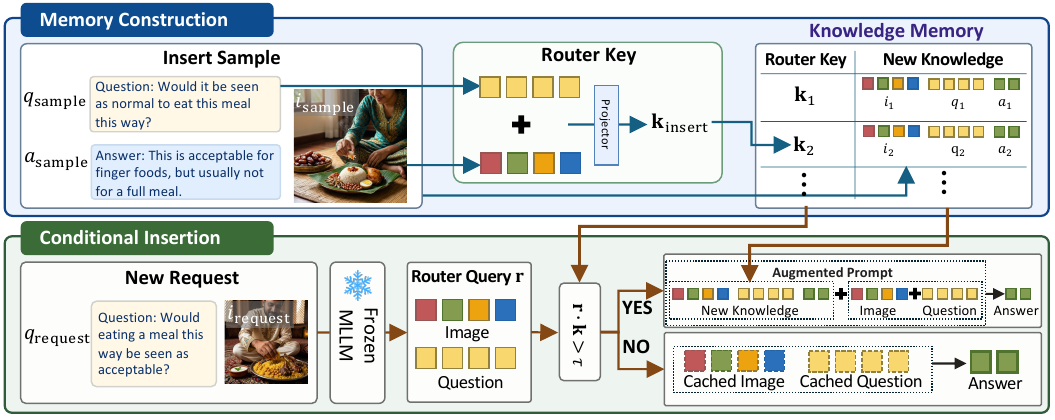}
\vspace{-3mm}
\caption{Overview of MCKI.
MCKI stores inserted knowledge in memory and uses a router to decide when to use it.}
\label{fig:method}
\vspace{-5mm}
\end{figure}

We instantiate Memory-Conditioned Knowledge Insertion~(MCKI) as a baseline for CrossCult-KIBench.
As shown in Fig.~\ref{fig:method}, MCKI stores each insertion sample in a knowledge memory and uses representations extracted by the frozen MLLM to decide whether a new request should activate a stored memory entry.
Only the lightweight router mapping is trained.

\subsection{Memory Construction}
For an insertion sample $e_j=(i_j,q_j,a_j)$, MCKI writes one entry $m_j=(\mathbf{k}_j,e_j)$ into the knowledge memory.
The entry stores a router key $\mathbf{k}_j$ together with the complete insertion sample $e_j$.
To build the router key $\mathbf{k}_j$, MCKI encodes $(i_j,q_j)$ with the frozen MLLM and takes the hidden representations from its final language layer.
It mean-pools these representations over the question tokens and visual tokens, then applies a trainable router mapping $\mathbf{k}_j=\phi(\bar{\mathbf{h}}^{\mathrm{q}}_j,\bar{\mathbf{h}}^{\mathrm{v}}_j)$.

\subsection{Conditional Insertion}
For a new request $x=(i_{\mathrm{req}},q_{\mathrm{req}})$, MCKI performs the same last-layer feature extraction and applies the same router mapping to obtain the request vector $\mathbf{r}(x)=\phi(\bar{\mathbf{h}}^{\mathrm{q}}_x,\bar{\mathbf{h}}^{\mathrm{v}}_x)$.
The router scores each memory entry by cosine similarity between the request vector and the memory key, and selects the entry with the largest score,
\begin{align}
\operatorname{sim}(x,m_j) &= \frac{\mathbf{r}(x)^\top \mathbf{k}_j}{\|\mathbf{r}(x)\|_2\|\mathbf{k}_j\|_2},\quad j^* = \arg\max_j \operatorname{sim}(x,m_j).
\end{align}
When the selected similarity falls below the calibrated threshold, MCKI reuses the cached base state and continues decoding with the frozen MLLM.
At or above the threshold, MCKI prepends the selected memory entry to the new request and reruns generation with the augmented prompt.
Formally, MCKI produces the output $y$ as
\begin{align}
\tilde{x}_{j^*}
&=
\bigl[\mathrm{Wrap}(e_{j^*}),\,i_{\mathrm{req}},\,q_{\mathrm{req}}\bigr],\\
y
&=
\begin{cases}
f(x), & \operatorname{sim}(x,m_{j^*})< \tau, \\
f(\tilde{x}_{j^*}), & \operatorname{sim}(x,m_{j^*})\ge \tau .
\end{cases}
\end{align}
Here, $\tau$ is the calibrated threshold, $\tilde{x}_{j^*}$ is the augmented request, and $\mathrm{Wrap}(\cdot)$ formats the selected insertion sample $e_{j^*}$ as reference context.
The router is trained with a contrastive loss that increases router scores for insertion samples and paired Generality items relative to Locality items, and $\tau$ is calibrated by maximizing routing accuracy on positive and negative training scores, as detailed in Appendix~\ref{app:mcki_training}.

\section{Experiments}

\begin{table*}[t]
\centering
\caption{Single-insert knowledge insertion results.
ROUGE-L scores are reported as percentages, and LLM Judge scores are reported on a 0 to 10 scale.
Higher scores indicate better performance.}
\label{tab:ki-single}
\begingroup
\setlength{\tabcolsep}{4pt}
\resizebox{\textwidth}{!}{%
\begin{tabular}{lcccccccccc}
\specialrule{1.2pt}{0pt}{0pt}
Method & \multicolumn{2}{c}{Reliability} & \multicolumn{2}{c}{Generality} & \multicolumn{2}{c}{Cross-Lang. Loc.} & \multicolumn{2}{c}{Cross-Scen. Loc.} & \multicolumn{2}{c}{Overall} \\ \cline{2-11}
 & ROUGE-L & LLM Judge & ROUGE-L & LLM Judge & ROUGE-L & LLM Judge & ROUGE-L & LLM Judge & ROUGE-L & LLM Judge \\ \hline
\multicolumn{11}{l}{\textbf{Base Model: InternVL3.5-8B}} \\ \hline
Base Model & 12.24 & 4.02 & 12.23 & 4.03 & 100.00 & 9.99 & 100.00 & 9.98 & 56.12 & 7.00 \\
\cdashline{1-11}[0.6pt/2pt]
FineTune & \textbf{91.12} & \textbf{9.35} & \textbf{24.66} & \textbf{6.32} & 5.36 & 2.28 & 4.96 & 0.90 & 31.53 & 4.71 \\
IKE & 70.25 & 8.54 & \underline{23.58} & \underline{6.28} & 24.92 & 5.14 & 34.46 & 6.45 & 38.30 & 6.60 \\
MEND & 19.24 & 4.82 & 19.33 & 4.87 & 67.16 & 8.58 & 42.59 & 6.89 & 37.08 & 6.29 \\
SERAC & 57.69 & 7.64 & 22.56 & 5.63 & \underline{78.81} & 9.07 & 26.51 & 4.38 & 46.39 & 6.68 \\
MSCKE & 69.93 & 8.48 & 23.55 & 5.80 & \textbf{100.00} & \underline{9.99} & \textbf{91.55} & \underline{9.38} & \underline{71.26} & \underline{8.41} \\
MCKI~(Ours) & \underline{79.83} & \underline{8.86} & 22.20 & 5.71 & \textbf{100.00} & \textbf{9.99} & \underline{89.66} & \textbf{9.47} & \textbf{72.92} & \textbf{8.51} \\ \hline
\multicolumn{11}{l}{\textbf{Base Model: Qwen3.5-9B}} \\ \hline
Base Model & 14.13 & 4.50 & 14.06 & 4.43 & 100.00 & 10.00 & 100.00 & 10.00 & 57.05 & 7.23 \\
\cdashline{1-11}[0.6pt/2pt]
FineTune & 55.90 & 6.35 & 15.54 & 4.37 & 0.00 & 0.96 & 3.17 & 0.61 & 18.65 & 3.07 \\
IKE & 47.25 & 8.06 & 23.76 & \textbf{6.84} & 19.93 & 5.20 & 35.72 & 5.63 & 31.66 & 6.43 \\
MEND & 44.91 & 6.52 & 26.31 & 5.80 & \underline{81.83} & 9.22 & \underline{61.32} & \underline{8.03} & 53.59 & 7.39 \\
SERAC & \underline{67.59} & \underline{8.23} & \textbf{27.59} & 6.42 & 5.75 & 3.98 & 18.88 & 3.92 & 29.95 & 5.64 \\
MSCKE & \textbf{71.27} & \textbf{8.50} & \underline{27.58} & 6.27 & \textbf{100.00} & \underline{10.00} & 28.64 & 4.94 & \underline{56.87} & \underline{7.43} \\
MCKI~(Ours) & 50.66 & 7.77 & 24.43 & \underline{6.74} & \textbf{100.00} & \textbf{10.00} & \textbf{88.27} & \textbf{9.32} & \textbf{65.84} & \textbf{8.46} \\ \specialrule{1.2pt}{0pt}{0pt}
\end{tabular}%
}
\endgroup
\vspace{-5mm}
\end{table*}

\begin{table*}[t]
\centering
\caption{Sequential-insert knowledge insertion results.}
\label{tab:ki-sequential}
\begingroup
\setlength{\tabcolsep}{4pt}
\resizebox{0.85\textwidth}{!}{%
\begin{tabular}{lcccccccc}
\specialrule{1.2pt}{0pt}{0pt}
Method & \multicolumn{2}{c}{Final Rel.} & \multicolumn{2}{c}{Final Gen.} & \multicolumn{2}{c}{Final Loc.} & \multicolumn{2}{c}{Overall} \\ \cline{2-9}
 & ROUGE-L & LLM Judge & ROUGE-L & LLM Judge & ROUGE-L & LLM Judge & ROUGE-L & LLM Judge \\ \hline
\multicolumn{9}{l}{\textbf{Base Model: InternVL3.5-8B}} \\ \hline
Base Model & 13.55 & 4.49 & 13.53 & 4.48 & 100.00 & 9.89 & 42.36 & 6.29 \\
\cdashline{1-9}[0.6pt/2pt]
FineTune & 6.80 & 0.90 & 1.85 & 0.60 & 0.26 & 0.12 & 2.97 & 0.54 \\
IKE & \underline{75.44} & 8.49 & \underline{27.44} & \textbf{6.46} & 28.87 & 5.99 & 43.92 & 6.98 \\
MEND & 2.42 & 0.74 & 2.36 & 0.74 & 3.19 & 0.93 & 2.66 & 0.80 \\
SERAC & 46.62 & 7.24 & 23.92 & 6.01 & 25.37 & 4.84 & 31.97 & 6.03 \\
MSCKE & 66.08 & \underline{8.80} & 23.48 & \underline{6.37} & \underline{56.28} & \underline{6.96} & \underline{48.61} & \underline{7.38} \\
MCKI~(Ours) & \textbf{85.57} & \textbf{8.98} & \textbf{28.61} & 6.26 & \textbf{83.44} & \textbf{9.18} & \textbf{65.87} & \textbf{8.14} \\ \hline
\multicolumn{9}{l}{\textbf{Base Model: Qwen3.5-9B}} \\ \hline
Base Model & 12.10 & 4.44 & 12.07 & 4.43 & 100.00 & 10.00 & 41.39 & 6.29 \\
\cdashline{1-9}[0.6pt/2pt]
FineTune & 2.37 & 0.44 & 1.00 & 0.32 & 0.28 & 0.10 & 1.22 & 0.29 \\
IKE & 13.85 & 2.22 & 7.01 & 1.87 & \textbf{78.98} & \underline{8.64} & 33.28 & 4.24 \\
MEND & 4.07 & 0.77 & 3.91 & 0.86 & 4.17 & 1.26 & 4.05 & 0.96 \\
SERAC & 36.23 & 6.04 & 19.09 & 5.02 & 14.23 & 4.14 & 23.18 & 5.07 \\
MSCKE & \textbf{62.10} & \textbf{8.54} & \underline{28.32} & \underline{6.49} & 42.77 & 6.18 & \underline{44.40} & \underline{7.07} \\
MCKI~(Ours) & \underline{59.83} & \underline{8.38} & \textbf{30.07} & \textbf{7.19} & \underline{73.68} & \textbf{8.89} & \textbf{54.53} & \textbf{8.15} \\ \specialrule{1.2pt}{0pt}{0pt}
\end{tabular}%
}
\endgroup
\vspace{-5mm}
\end{table*}

\subsection{Experimental Setup}

\textbf{Base MLLMs.}
We evaluate knowledge insertion methods on CrossCult-KIBench with two open-source MLLMs, InternVL3.5-8B~\citep{wang2025internvl3} and Qwen3.5-9B~\citep{qwen35blog}.
Both models provide strong visual question answering backbones while remaining accessible for controlled insertion experiments.

\textbf{Baseline Methods.}
Among existing approaches, knowledge editing methods are most closely related to our task, since they also aim to alter model knowledge or behavior. So we adapt representative editing methods as baselines for comparison with MCKI.
Following prior work on multimodal editing~\citep{mmedit,FGVEdit}, we include textual editing methods because the set of specialized multimodal editing methods remains limited.
\textbf{FineTune} updates the MLLM on the insertion sample with target answer supervision.
\textbf{IKE}~\citep{ike} uses the insertion sample as an in-context demonstration at inference time.
\textbf{MEND}~\citep{mend} applies a learned gradient-based edit from the target answer loss.
\textbf{SERAC}~\citep{serac} retrieves relevant edits and answers with a replacement model when an edit is activated.
\textbf{MSCKE}~\citep{FGVEdit} uses a multimodal CLIP router and a replacement model to apply inserted knowledge.

\subsection{Main Results}

\paragraph{Single-Insert Results.}
Table~\ref{tab:ki-single} shows results on the single-insert task.
Direct target supervision gives FineTune high Reliability, yet the same parameter updates overwrite existing behavior so aggressively that both Cross-Language Locality and Cross-Scenario Locality suffer severe drops.
IKE avoids weight-level damage by prepending the insertion sample as an in-context demonstration at inference time, but its effectiveness depends on whether the base model can use that demonstration, which leaves its Reliability and Overall improvements limited.
Among learned editing methods, MEND applies more conservative parameter changes and retains more locality than FineTune, but this conservatism also weakens target fitting.
Retrieval-based replacement methods expose a different trade-off.
SERAC improves Reliability and Generality when the edit is retrieved, yet similar non-target evaluation requests can also activate the replacement path and erode locality.
MSCKE adds multimodal routing and better preserves Cross-Language Locality, but related scenes can still activate replacement behavior and weaken Cross-Scenario Locality.
MCKI shows that selective memory activation can yield a more balanced Overall profile by preserving cross-language behavior while maintaining competitive target scores.
No existing method balances all four dimensions, showing that CrossCult-KIBench exposes complementary failure modes in cross-cultural knowledge insertion.

\paragraph{Sequential-Insert Results.}
Table~\ref{tab:ki-sequential} shows that sequential insertion adds challenges that are absent from single insertion.
After several insertions, methods must keep multiple culture-conditioned answers available while preserving unrelated behavior, so failures reflect overwriting, model state drift, and retrieval confusion rather than target fitting alone.
FineTune and MEND are most affected because repeated parameter updates can overwrite earlier behavior.
IKE avoids weight changes, but its final performance depends on whether the model can use multiple demonstrations without confusing or ignoring earlier insertions.
SERAC and MSCKE benefit from explicit memory or routing and therefore maintain stronger target scores than collapsed update methods, yet larger memories and related image-language contexts can still trigger the wrong replacement path and reduce Final Locality.
MCKI gives a stronger Overall balance because selective memory activation helps preserve unrelated behavior while retaining inserted answers, but its lower Final Generality shows that storing sequential insertions does not by itself transfer them to related examples from the same scenario.
No existing method fully resolves coexistence under repeated insertion, which highlights cross-cultural knowledge insertion as a task with substantial room for further exploration.

\begin{figure*}[t]
\centering
\includegraphics[width=\textwidth]{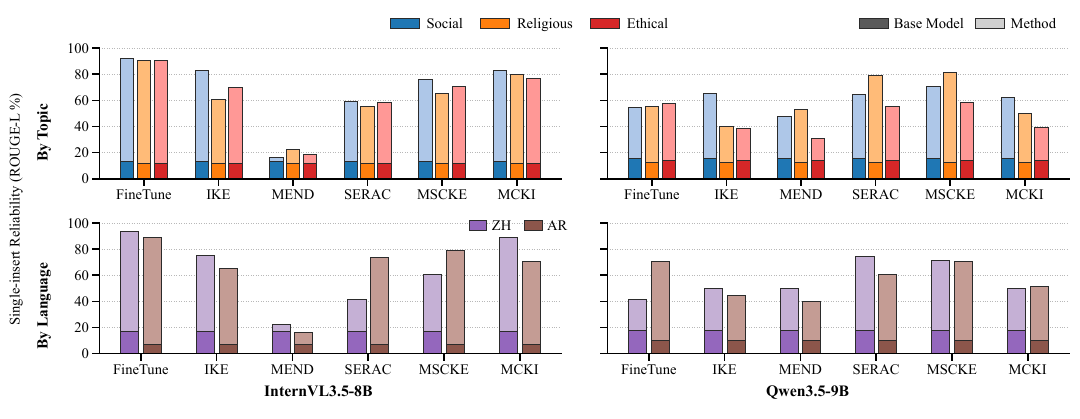}
\vspace{-5mm}
\caption{Single-insert reliability by topic group and inserted language-culture partition.
Dark bars show the base model and light bars show the edited model.}
\label{fig:topic_partition_gain}
\vspace{-3mm}
\end{figure*}

\subsection{Topic and Partition Sensitivity}

To examine whether insertion difficulty depends on cultural topic or language-culture partition, we analyze single-insert Reliability by topic group and inserted partition in Fig.~\ref{fig:topic_partition_gain}.
Across topic groups, the two backbones show different patterns.
InternVL3.5-8B is relatively stable with a mild drop on Religious cases, whereas Qwen3.5-9B shows weaker gains on Ethical cases.
This suggests that topic difficulty reflects how a backbone represents cultural cues, since Religious cases often contain more explicit symbolic triggers while Social and Ethical cases require broader semantic transfer.
Partition effects are also model-dependent, with Arabic insertions generally leaving more room for Reliability improvement than Chinese insertions but with a gap that varies across backbones and insertion mechanisms.

\subsection{Sequential-Insert Retention Analysis}
\begin{figure*}[t]
\centering
\includegraphics[width=\textwidth]{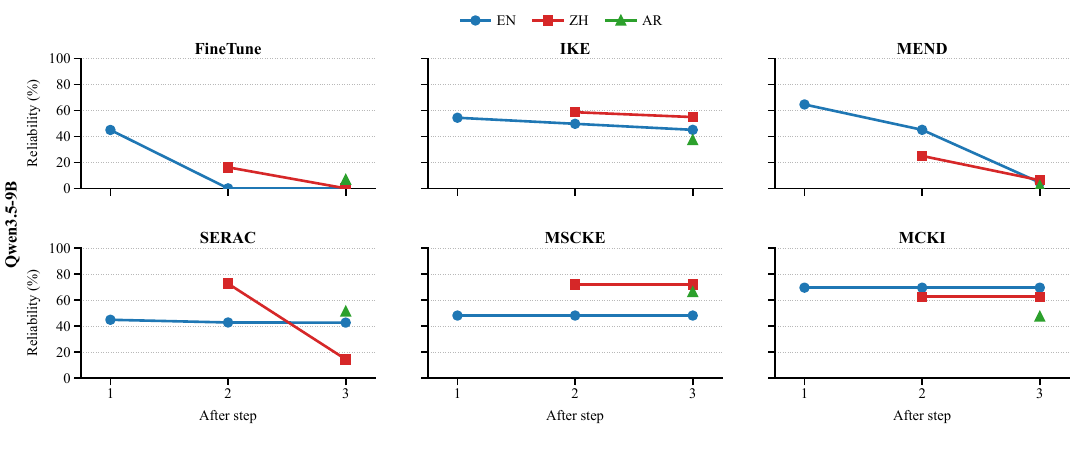}
\vspace{-10mm}
\caption{Reliability trajectories after sequential insertion on Qwen3.5-9B.}
\label{fig:sequential_retention}
\vspace{-3mm}
\end{figure*}

We also evaluated the impact of each insertion on the knowledge inserted before.
Fig.~\ref{fig:sequential_retention} shows the results.
IKE retains earlier insertions more steadily than methods that update parameters, while FineTune and MEND rapidly lose them after later steps.
Retrieval-based and routing-based methods retain inserted answers more consistently, with MSCKE and MCKI giving the strongest retention among the evaluated methods.
Additional retention trajectories on InternVL3.5-8B and reordered insertion chains are provided in Appendix~\ref{app:sequential_retention}.

\subsection{Efficiency Analysis}

\begin{table*}[t]
\centering
\caption{Efficiency Analysis on 100 sampled training cases and 100 sampled single-insert evaluation cases.
Parameters count the total loaded parameters.}
\label{tab:runtime-efficiency}
\begingroup
\setlength{\tabcolsep}{3.2pt}
\resizebox{0.8\textwidth}{!}{%
\begin{tabular}{llrrrrrr}
\specialrule{1.2pt}{0pt}{0pt}
Backbone & Method & Train / case & Insert / case & Request / sample & Train Mem. & Eval Mem. & Params \\
 & & ms & ms & ms & GiB & GiB & B \\ \hline
\multirow{6}{*}{InternVL3.5-8B}
& FineTune & - & 13954.3 & 2002.0 & - & 73.6 & 8.53 \\
& IKE & - & 4.5 & 1581.4 & - & 24.3 & 8.55 \\
& MEND & 5251.5 & 867.7 & 1822.1 & 44.1 & 71.3 & 8.53 \\
& SERAC & 2017.7 & 4.6 & 1325.1 & 54.9 & 27.7 & 9.61 \\
& MSCKE & 361.9 & 48.3 & 1333.4 & 28.0 & 29.2 & 10.02 \\
& MCKI & 1655.3 & 306.0 & 1902.0 & 17.7 & 21.1 & 8.54 \\ \hline
\multirow{6}{*}{Qwen3.5-9B}
& FineTune & - & 5700.6 & 1981.1 & - & 41.9 & 9.41 \\
& IKE & - & 4.5 & 1231.6 & - & 22.1 & 9.43 \\
& MEND & 2898.8 & 318.3 & 1430.0 & 25.7 & 39.3 & 9.41 \\
& SERAC & 1211.5 & 4.7 & 735.6 & 22.8 & 20.1 & 10.29 \\
& MSCKE & 308.9 & 45.8 & 862.2 & 22.3 & 21.6 & 10.69 \\
& MCKI & 1026.6 & 107.8 & 943.9 & 17.8 & 18.0 & 9.42 \\ \specialrule{1.2pt}{0pt}{0pt}
\end{tabular}%
}
\endgroup
\vspace{-3mm}
\end{table*}

To analyze insertion efficiency across methods, we measure runtime cost on one NVIDIA A100-SXM4 80GB GPU using 100 training cases and 100 single-insert evaluation cases.
Table~\ref{tab:runtime-efficiency} reports measured latency, peak GPU memory, and loaded parameter count.
Request time includes the full routing or replacement path and autoregressive generation.
FineTune requires no training, but it incurs highest insertion latency and the largest evaluation memory.
MEND is also expensive because each edit requires gradient-based parameter updates and high memory during both training and evaluation.
IKE and SERAC have very low insertion latency, but their request latency depends on in-context prompting or replacement generation, and SERAC loads an additional replacement model.
MSCKE has moderate insertion cost and low request latency, but its replacement model increases the total loaded parameter count.
MCKI keeps the loaded parameter count close to the base MLLM and uses substantially less evaluation memory than FineTune and MEND, while adding moderate router training and insertion cost.

\section{Conclusion}
In this work, we introduce the task of cross-cultural knowledge insertion: the post-hoc augmentation of MLLMs with missing culture-specific knowledge, while preserving their original behavior for all other cultural contexts. This formulation enables targeted cultural adaptation without full retraining.
To enable rigorous evaluation, we present CrossCult-KIBench, a benchmark built on 49 visually grounded cultural scenarios across English, Chinese, and Arabic language-culture partitions. It supports both single-insert and sequential-insert evaluation, measuring not only insertion success, but also cross-cultural transfer, behavioral preservation, and stability across multiple updates.
Experiments reveal that existing knowledge-editing and retrieval-based approaches struggle to balance these objectives, often improving target culture performance at the expense of non-target behavior. As a strong and practical baseline, we propose MCKI, which offers superior performance across multiple metrics.
Together, CrossCult-KIBench and the MCKI baseline establish a foundation for developing and evaluating more precise, effective, and culturally-aware knowledge insertion methods for MLLMs.

{\small
\bibliographystyle{IEEEtran}
\bibliography{references}

@String(ICCV= {Int. Conf. Comput. Vis.})

@String(ECCV= {Eur. Conf. Comput. Vis.})

@String(ICLR = {Int. Conf. Learn. Represent.})

@String(ICCV  = {ICCV})

@String(ECCV  = {ECCV})

@String(ICLR  = {ICLR})

@article{wang2024vlbiasbench,
  title={Vlbiasbench: A comprehensive benchmark for evaluating bias in large vision-language model},
  author={Wang, Sibo and Cao, Xiangkui and Zhang, Jie and Yuan, Zheng and Shan, Shiguang and Chen, Xilin and Gao, Wen},
  journal={arXiv preprint arXiv:2406.14194},
  year={2024}
}

@article{romero2024cvqa, 
  title={Cvqa: Culturally-diverse multilingual visual question answering benchmark},
  author={Romero, David and Lyu, Chenyang and Wibowo, Haryo Akbarianto and Lynn, Teresa and Hamed, Injy and Kishore, Aditya Nanda and Mandal, Aishik and Dragonetti, Alina and Abzaliev, Artem and Tonja, Atnafu Lambebo and others},
  journal={arXiv preprint arXiv:2406.05967},
  year={2024}
}

@inproceedings{liu2021visually, 
  title={Visually grounded reasoning across languages and cultures},
  author={Liu, Fangyu and Bugliarello, Emanuele and Ponti, Edoardo Maria and Reddy, Siva and Collier, Nigel and Elliott, Desmond},
  booktitle={Proceedings of the 2021 Conference on Empirical Methods in Natural Language Processing},
  pages={10467--10485},
  year={2021}
}

@article{li2024culturepark, 
  title={Culturepark: Boosting cross-cultural understanding in large language models},
  author={Li, Cheng and Teney, Damien and Yang, Linyi and Wen, Qingsong and Xie, Xing and Wang, Jindong},
  journal={Advances in Neural Information Processing Systems},
  volume={37},
  pages={65183--65216},
  year={2024}
}

@article{li2024culturellm, 
  title={Culturellm: Incorporating cultural differences into large language models},
  author={Li, Cheng and Chen, Mengzhuo and Wang, Jindong and Sitaram, Sunayana and Xie, Xing},
  journal={Advances in Neural Information Processing Systems},
  volume={37},
  pages={84799--84838},
  year={2024}
}

@article{chen2017chinesefoodnet, 
  title={Chinesefoodnet: A large-scale image dataset for chinese food recognition},
  author={Chen, Xin and Zhu, Yu and Zhou, Hua and Diao, Liang and Wang, Dongyan},
  journal={arXiv preprint arXiv:1705.02743},
  year={2017}
}

@InProceedings{Kapitanov_2024_WACV,
    author    = {Kapitanov, Alexander and Kvanchiani, Karina and Nagaev, Alexander and Kraynov, Roman and Makhliarchuk, Andrei},
    title     = {HaGRID -- HAnd Gesture Recognition Image Dataset},
    booktitle = {Proceedings of the IEEE/CVF Winter Conference on Applications of Computer Vision (WACV)},
    month     = {January},
    year      = {2024},
    pages     = {4572-4581}
}

@inproceedings{jia2020fashionpedia, 
  title={Fashionpedia: Ontology, Segmentation, and an Attribute Localization Dataset},
  author={Jia, Menglin and Shi, Mengyun and Sirotenko, Mikhail and Cui, Yin and Cardie, Claire and Hariharan, Bharath and Adam, Hartwig and Belongie, Serge},
  booktitle={European Conference on Computer Vision (ECCV)},
  year={2020}
}

@inproceedings{VireoFood172,
  title={Deep-based ingredient recognition for cooking recipe retrieval},
  author={Chen, Jingjing and Ngo, Chong-Wah},
  booktitle={Proceedings of the 24th ACM international conference on Multimedia},
  pages={32--41},
  year={2016}
}

@inproceedings{netzer2011reading, 
  title={Reading digits in natural images with unsupervised feature learning},
  author={Netzer, Yuval and Wang, Tao and Coates, Adam and Bissacco, Alessandro and Wu, Baolin and Ng, Andrew Y and others},
  booktitle={NIPS workshop on deep learning and unsupervised feature learning},
  volume={2011},
  number={2},
  pages={4},
  year={2011},
  organization={Granada}
}

@inproceedings{xu2018towards, 
  title={Towards End-to-End License Plate Detection and Recognition: A Large Dataset and Baseline},
  author={Xu, Zhenbo and Yang, Wei and Meng, Ajin and Lu, Nanxue and Huang, Huan},
  booktitle={Proceedings of the European Conference on Computer Vision (ECCV)},
  pages={255--271},
  year={2018}
}

@article{kuznetsova2020open, 
  title={The open images dataset v4: Unified image classification, object detection, and visual relationship detection at scale},
  author={Kuznetsova, Alina and Rom, Hassan and Alldrin, Neil and Uijlings, Jasper and Krasin, Ivan and Pont-Tuset, Jordi and Kamali, Shahab and Popov, Stefan and Malloci, Matteo and Kolesnikov, Alexander and others},
  journal={International journal of computer vision},
  volume={128},
  number={7},
  pages={1956--1981},
  year={2020},
  publisher={Springer}
}

@article{touvron2023llama,
  title={Llama: Open and efficient foundation language models},
  author={Touvron, Hugo and Lavril, Thibaut and Izacard, Gautier and Martinet, Xavier and Lachaux, Marie-Anne and Lacroix, Timoth{\'e}e and Rozi{\`e}re, Baptiste and Goyal, Naman and Hambro, Eric and Azhar, Faisal and others},
  journal={arXiv preprint arXiv:2302.13971},
  year={2023}
}

@article{zhao2023survey,
  title={A survey of large language models},
  author={Zhao, Wayne Xin and Zhou, Kun and Li, Junyi and Tang, Tianyi and Wang, Xiaolei and Hou, Yupeng and Min, Yingqian and Zhang, Beichen and Zhang, Junjie and Dong, Zican and others},
  journal={arXiv preprint arXiv:2303.18223},
  year={2023}
}

@inproceedings{mmedit,
    title = "Can We Edit Multimodal Large Language Models?",
    author = "Cheng, Siyuan  and
      Tian, Bozhong  and
      Liu, Qingbin  and
      Chen, Xi  and
      Wang, Yongheng  and
      Chen, Huajun  and
      Zhang, Ningyu",
    booktitle = "EMNLP",
    year = "2023"
}

@inproceedings{mend,
title={Fast Model Editing at Scale},
author={Eric Mitchell and Charles Lin and Antoine Bosselut and Chelsea Finn and Christopher D Manning},
booktitle={ICLR},
year={2022}
}

@inproceedings{serac,
  title={Memory-based model editing at scale},
  author={Mitchell, Eric and Lin, Charles and Bosselut, Antoine and Manning, Christopher D and Finn, Chelsea},
  booktitle={ICML},
  year={2022}
}

@inproceedings{ike,
  title={Can We Edit Factual Knowledge by In-Context Learning?},
  author={Zheng, Ce and Li, Lei and Dong, Qingxiu and Fan, Yuxuan and Wu, Zhiyong and Xu, Jingjing and Chang, Baobao},
  booktitle={EMNLP},
  year={2023}
}

@article{kebench,
  title={Vlkeb: A large vision-language model knowledge editing benchmark},
  author={Huang, Han and Zhong, Haitian and Yu, Tao and Liu, Qiang and Wu, Shu and Wang, Liang and Tan, Tieniu},
  journal={Advances in Neural Information Processing Systems},
  volume={37},
  pages={9257--9280},
  year={2024}
}

@article{mike,
  title={MIKE: A New Benchmark for Fine-grained Multimodal Entity Knowledge Editing},
  author={Li, Jiaqi and Du, Miaozeng and Zhang, Chuanyi and Chen, Yongrui and Hu, Nan and Qi, Guilin and Jiang, Haiyun and Cheng, Siyuan and Tian, Bozhong},
  journal={arXiv preprint arXiv:2402.14835},
  year={2024}
}

@article{mcmke,
  title={MC-MKE: A Fine-Grained Multimodal Knowledge Editing Benchmark Emphasizing Modality Consistency},
  author={Zhang, Junzhe and Zhang, Huixuan and Yin, Xunjian and Huang, Baizhou and Zhang, Xu and Hu, Xinyu and Wan, Xiaojun},
  journal={arXiv preprint arXiv:2406.13219},
  year={2024}
}

@inproceedings{yao2023editing,
title={Editing Large Language Models: Problems, Methods, and Opportunities},
author={Yunzhi Yao and Peng Wang and Bozhong Tian and Siyuan Cheng and Zhoubo Li and Shumin Deng and Huajun Chen and Ningyu Zhang},
booktitle={EMNLP},
year={2023}
}

@InProceedings{FGVEdit,
    author    = {Zeng, Zhen and Gu, Leijiang and Yang, Xun and Duan, Zhangling and Shi, Zenglin and Wang, Meng},
    title     = {Visual-Oriented Fine-Grained Knowledge Editing for MultiModal Large Language Models},
    booktitle = {Proceedings of the IEEE/CVF International Conference on Computer Vision (ICCV)},
    month     = {October},
    year      = {2025},
    pages     = {2491-2500}
}

@article{lin2024mala,
  title={Mala-500: Massive language adaptation of large language models},
  author={Lin, Peiqin and Ji, Shaoxiong and Tiedemann, J{\"o}rg and Martins, Andr{\'e} FT and Sch{\"u}tze, Hinrich},
  journal={arXiv preprint arXiv:2401.13303},
  year={2024}
}

@inproceedings{nguyen2024seallms,
  title={SeaLLMs-large language models for Southeast Asia},
  author={Nguyen, Xuan-Phi and Zhang, Wenxuan and Li, Xin and Aljunied, Mahani and Hu, Zhiqiang and Shen, Chenhui and Chia, Yew Ken and Li, Xingxuan and Wang, Jianyu and Tan, Qingyu and others},
  booktitle={Proceedings of the 62nd Annual Meeting of the Association for Computational Linguistics (Volume 3: System Demonstrations)},
  pages={294--304},
  year={2024}
}

@misc{qwen35blog,
    title = {Qwen3.5: Accelerating Productivity with Native Multimodal Agents},
    url = {https://qwen.ai/blog?id=qwen3.5},
    author = {Qwen Team},
    month = {February},
    year = {2026}
}

@article{wang2025internvl3,
  title={Internvl3. 5: Advancing open-source multimodal models in versatility, reasoning, and efficiency},
  author={Wang, Weiyun and Gao, Zhangwei and Gu, Lixin and Pu, Hengjun and Cui, Long and Wei, Xingguang and Liu, Zhaoyang and Jing, Linglin and Ye, Shenglong and Shao, Jie and others},
  journal={arXiv preprint arXiv:2508.18265},
  year={2025}
}

@misc{survey2022,
  author = {{World Values Survey}},
  title = {World Values Survey},
  year = {2022},
  howpublished = {\url{https://www.worldvaluessurvey.org/wvs.jsp}}
}

@article{bai2025qwen3,
  title={Qwen3-vl technical report},
  author={Bai, Shuai and Cai, Yuxuan and Chen, Ruizhe and Chen, Keqin and Chen, Xionghui and Cheng, Zesen and Deng, Lianghao and Ding, Wei and Gao, Chang and Ge, Chunjiang and others},
  journal={arXiv preprint arXiv:2511.21631},
  year={2025}
}

@article{gemmateam2025gemma3technicalreport,
      title={Gemma 3 Technical Report}, 
      author={{Gemma Team} and Aishwarya Kamath and Johan Ferret and Shreya Pathak and Nino Vieillard and Ramona Merhej and Sarah Perrin and Tatiana Matejovicova and Alexandre Ramé and others},
      journal={arXiv preprint arXiv:2503.19786},
      year={2025},
}

@article{zhang2024mm,
  title={Mm-llms: Recent advances in multimodal large language models},
  author={Zhang, Duzhen and Yu, Yahan and Dong, Jiahua and Li, Chenxing and Su, Dan and Chu, Chenhui and Yu, Dong},
  journal={Findings of the Association for Computational Linguistics: ACL 2024},
  pages={12401--12430},
  year={2024}
}

@article{huang2024survey,
  title={A survey on evaluation of multimodal large language models},
  author={Huang, Jiaxing and Zhang, Jingyi},
  journal={arXiv preprint arXiv:2408.15769},
  year={2024}
}

@article{wu2025qwen,
  title={Qwen-image technical report},
  author={Wu, Chenfei and Li, Jiahao and Zhou, Jingren and Lin, Junyang and Gao, Kaiyuan and Yan, Kun and Yin, Sheng-ming and Bai, Shuai and Xu, Xiao and Chen, Yilei and others},
  journal={arXiv preprint arXiv:2508.02324},
  year={2025}
}
}

\clearpage

\appendix

\section{Limitations}
\label{app:limitations}

We acknowledge a limitation of CrossCult-KIBench that suggests a promising direction for future research.
The main limitation is that the benchmark uses language-culture partitions as practical proxies for cultural settings.
Specifically, English, Chinese, and Arabic are paired with the U.S. context, China context, and Arab region context to make insertion targets observable and evaluable within multilingual MLLMs.
This design makes the task operational, but it does not imply that a language uniquely defines a culture or that speakers of the same language share a single cultural norm.
Languages can span multiple regions, communities, religions, and social groups, and the same cultural practice can also appear across multiple languages.
As a result, CrossCult-KIBench should be interpreted as evaluating insertion under specified language-culture partitions rather than evaluating cultures in their full internal diversity.
This distinction is particularly relevant because our task is knowledge insertion rather than cultural inference.
The goal is not to infer a person's culture from language, nor to recommend culture-specific behavior for arbitrary users.
Instead, each insertion sample provides an explicit target setting and asks whether a method can insert that specified culture-conditioned response while preserving behavior outside the target partition.
In future work, we will extend this setting to finer cultural descriptors, multilingual variants within the same region, and cases where language and culture are decoupled, while preserving the insertion objective and its locality requirements.

\section{Broader Impacts}
\label{app:broader_impacts}

CrossCult-KIBench evaluates whether knowledge insertion methods can add specified culture-conditioned knowledge to image-grounded MLLMs while preserving behavior outside the target language-culture partition.
Its primary positive impact is to give researchers a structured way to audit insertion success and side effects in multilingual cultural settings.
The benchmark may also expose failures that are hidden by English-centered or culturally coarse evaluation, which can support more careful model assessment before downstream use.

The main risk is that language-culture partitions could be misread as fixed cultural identities or as prescriptions for individual users.
Another risk is that culture-conditioned insertion could be misused to steer model behavior toward stereotyped, manipulative, or politically sensitive outputs.
We mitigate these risks by framing CrossCult-KIBench as an evaluation setting with explicit target partitions, by distinguishing knowledge insertion from cultural inference, and by discussing the limits of language-culture proxies in Appendix~\ref{app:limitations}.

\section{Benchmark Details}
\label{app:benchmark_details}

\begin{figure}[h]
\centering
\includegraphics[width=0.98\linewidth]{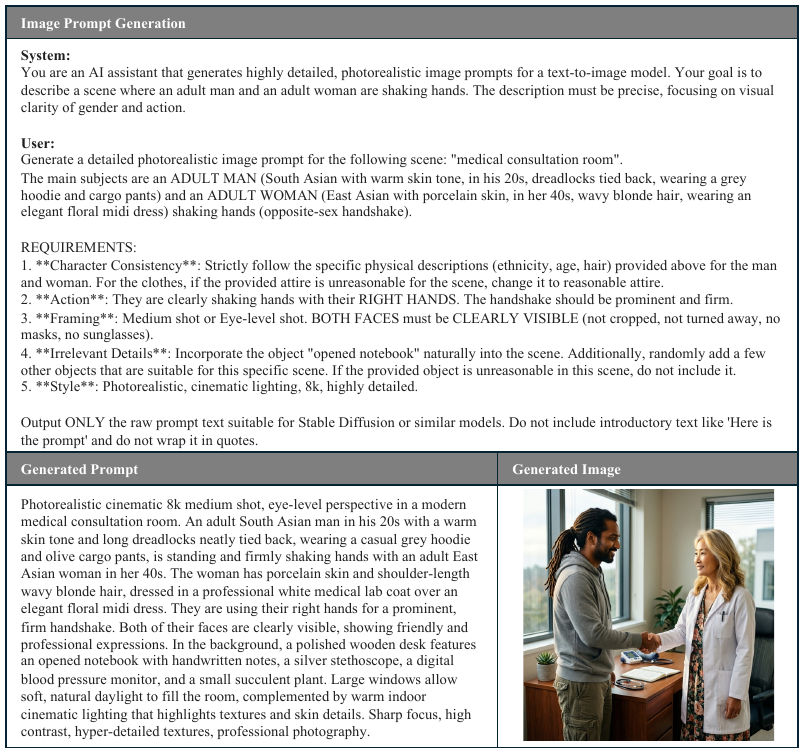}
\caption{Example of the LLM-assisted image prompt construction step.}
\label{fig:appendix_image_prompt}
\end{figure}

\begin{figure}[h]
\centering
\includegraphics[width=0.98\linewidth]{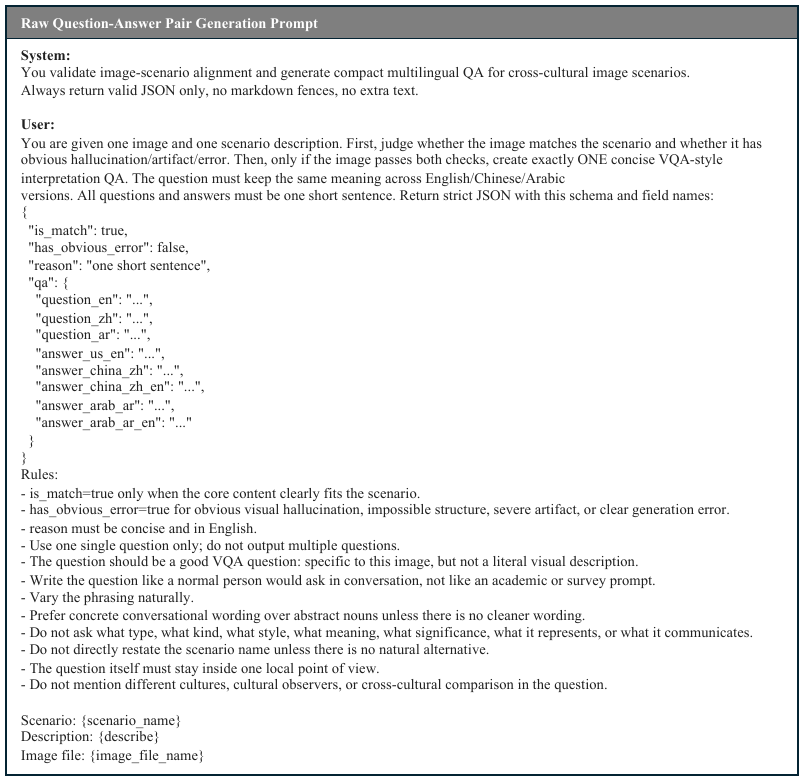}
\caption{Example of the LLM-assisted raw multilingual QA construction step.}
\label{fig:appendix_raw_qa_prompt}
\end{figure}

\subsection{Metric Details}
\label{app:metric_definitions}
\subsubsection{Single-Insert Metrics}
Let $\mathcal{C}_{\mathrm{single}}$ denote the set of single-insert cases.
For each $c \in \mathcal{C}_{\mathrm{single}}$, we write the insertion sample as $e_c=(i_c,q_c^{p_c},a_c^{p_c})$ and denote the post-insertion model by $f'_c$.
Reliability is
\begin{align}
\mathcal{M}_{\mathrm{reliability}}
&= \frac{1}{|\mathcal{C}_{\mathrm{single}}|}
\sum_{c \in \mathcal{C}_{\mathrm{single}}}
\mathrm{Score}\!\left(f'_c(i_c,q_c^{p_c}), a_c^{p_c}\right).
\end{align}
Let $e_{c,\mathrm{gen}}^{p_c}=(i_{c,\mathrm{gen}},q_{c,\mathrm{gen}}^{p_c},a_{c,\mathrm{gen}}^{p_c})$ denote the Generality evaluation item.
Generality is
\begin{align}
\mathcal{M}_{\mathrm{generality}}
&= \frac{1}{|\mathcal{C}_{\mathrm{single}}|}
\sum_{c \in \mathcal{C}_{\mathrm{single}}}
\mathrm{Score}\!\left(f'_c(i_{c,\mathrm{gen}},q_{c,\mathrm{gen}}^{p_c}), a_{c,\mathrm{gen}}^{p_c}\right).
\end{align}
Let $(i_c,q_{c,\mathrm{lang}}^{p_\ell})$ denote the Cross-Language Locality evaluation item, with $p_\ell \neq p_c$.
Cross-Language Locality is
\begin{align}
\mathcal{M}_{\mathrm{cross}\text{-}\mathrm{language}}
&= \frac{1}{|\mathcal{C}_{\mathrm{single}}|}
\sum_{c \in \mathcal{C}_{\mathrm{single}}}
\mathrm{Score}\!\left(f'_c(i_c,q_{c,\mathrm{lang}}^{p_\ell}), f(i_c,q_{c,\mathrm{lang}}^{p_\ell})\right).
\end{align}
Let $(i_{c,\mathrm{scen}},q_{c,\mathrm{scen}}^{p_c})$ denote the Cross-Scenario Locality evaluation item.
Cross-Scenario Locality is
\begin{align}
\mathcal{M}_{\mathrm{cross}\text{-}\mathrm{scenario}}
&= \frac{1}{|\mathcal{C}_{\mathrm{single}}|}
\sum_{c \in \mathcal{C}_{\mathrm{single}}}
\mathrm{Score}\!\left(f'_c(i_{c,\mathrm{scen}},q_{c,\mathrm{scen}}^{p_c}), f(i_{c,\mathrm{scen}},q_{c,\mathrm{scen}}^{p_c})\right).
\end{align}
Single-insert Overall is
\begin{align}
\mathcal{M}_{\mathrm{overall}}^{\mathrm{single}}
&= \frac{1}{4}
\left(
\mathcal{M}_{\mathrm{reliability}}
+ \mathcal{M}_{\mathrm{generality}}
+ \mathcal{M}_{\mathrm{cross}\text{-}\mathrm{language}}
+ \mathcal{M}_{\mathrm{cross}\text{-}\mathrm{scenario}}
\right).
\end{align}

\subsubsection{Sequential-Insert Metrics}
Let $\mathcal{C}_{\mathrm{seq}}$ denote the set of sequential-insert chains.
For each $c \in \mathcal{C}_{\mathrm{seq}}$, let $f'_c$ be the final post-insertion model after all three insertions.
Let $e_{c,t}^{p_{c,t}}=(i_c,q_{c,t}^{p_{c,t}},a_{c,t}^{p_{c,t}})$ denote the insertion sample for step $t$.
Final Reliability is
\begin{align}
\mathcal{M}_{\mathrm{reliability}}^{\mathrm{final}}
&= \frac{1}{3|\mathcal{C}_{\mathrm{seq}}|}
\sum_{c \in \mathcal{C}_{\mathrm{seq}}}
\sum_{t=1}^{3}
\mathrm{Score}\!\left(f'_c(i_c,q_{c,t}^{p_{c,t}}), a_{c,t}^{p_{c,t}}\right).
\end{align}
Let $e_{c,\mathrm{gen},t}^{p_{c,t}}=(i_{c,\mathrm{gen}},q_{c,\mathrm{gen},t}^{p_{c,t}},a_{c,\mathrm{gen},t}^{p_{c,t}})$ denote the Generality evaluation item for step $t$.
Final Generality is
\begin{align}
\mathcal{M}_{\mathrm{generality}}^{\mathrm{final}}
&= \frac{1}{3|\mathcal{C}_{\mathrm{seq}}|}
\sum_{c \in \mathcal{C}_{\mathrm{seq}}}
\sum_{t=1}^{3}
\mathrm{Score}\!\left(f'_c(i_{c,\mathrm{gen}},q_{c,\mathrm{gen},t}^{p_{c,t}}), a_{c,\mathrm{gen},t}^{p_{c,t}}\right).
\end{align}
Let $(i_{c,\mathrm{loc},t},q_{c,\mathrm{loc},t}^{p_{c,t}})$ denote the Locality evaluation item paired with step $t$.
Final Locality is
\begin{align}
\mathcal{M}_{\mathrm{locality}}^{\mathrm{final}}
&= \frac{1}{3|\mathcal{C}_{\mathrm{seq}}|}
\sum_{c \in \mathcal{C}_{\mathrm{seq}}}
\sum_{t=1}^{3}
\mathrm{Score}\!\left(f'_c(i_{c,\mathrm{loc},t},q_{c,\mathrm{loc},t}^{p_{c,t}}), f(i_{c,\mathrm{loc},t},q_{c,\mathrm{loc},t}^{p_{c,t}})\right).
\end{align}
Sequential-insert Overall is
\begin{align}
\mathcal{M}_{\mathrm{overall}}^{\mathrm{seq}}
&= \frac{1}{3}
\left(
\mathcal{M}_{\mathrm{reliability}}^{\mathrm{final}}
+ \mathcal{M}_{\mathrm{generality}}^{\mathrm{final}}
+ \mathcal{M}_{\mathrm{locality}}^{\mathrm{final}}
\right).
\end{align}

\subsection{Benchmark Construction Details}
\label{app:benchmark_construction_details}

CrossCult-KIBench is built through a human-led pipeline with constrained model assistance.
We first define the three topic groups, identify and verify public cultural sources, and write scenario metadata that specifies the visual situation, the partition-specific cultural context, the expected judgment boundary, question-style examples, question design notes, and image-check notes.
The metadata then controls image preparation, where candidate images are collected from public image sources when suitable candidates are available and generated otherwise.
For generated-image scenarios, gpt-5.4-mini expands the scenario configuration into a detailed image-generation prompt, as illustrated in Fig.~\ref{fig:appendix_image_prompt} for a handshake scenario involving a man and a woman.
Qwen-Image~\citep{wu2025qwen} uses the resulting positive prompt together with the scenario negative prompt and suffix to generate candidate images.
We then manually screen all collected and generated candidates for scenario-image alignment, visual artifacts, and suitability.
After the verified image set is finalized, Gemini-3.1-Flash-Lite drafts raw multilingual QA items from each image and its scenario metadata, as shown in Fig.~\ref{fig:appendix_raw_qa_prompt}.
This prompt first instructs the model to check whether the image matches the scenario and whether it contains obvious generation errors.
Only when the image passes both checks does the model produce one QA item with aligned English, Chinese, and Arabic question versions and partition-specific answers.
The accepted raw multilingual items are finally converted into benchmark cases.
For single-insert evaluation, the Chinese and Arabic partitions become target insertion samples.
For sequential-insert evaluation, the English, Chinese, and Arabic samples from the same raw multilingual case form a three-step insertion chain.

\section{MCKI Details}
\label{app:mcki_details}

\subsection{Training of MCKI}
\label{app:mcki_training}

MCKI trains a lightweight router while keeping the base MLLM frozen.
The router maps frozen hidden states to the memory key and request vector.
These modules include the text and visual normalization layers, the text and visual projection layers, and the fusion projection that produces the final normalized router vector.
The stored insertion sample is not optimized.
It serves as the reference payload when a memory entry is activated.

For memory entry $m_j$, let $\mathcal{P}_j$ contain the insertion sample $e_j$ and its paired Generality evaluation items.
Let $\mathcal{N}_j$ contain the paired Locality evaluation items.
The positive set trains the router to activate the same memory entry for the inserted case and its intended Generality variants.
The negative set trains the router to abstain when the base behavior should be preserved.
With logit scale $\gamma$, define the positive and negative logits as
\begin{align}
\ell_p^+
&=
\gamma\operatorname{sim}(p,m_j),\\
\ell_n^-
&=
\gamma\operatorname{sim}(n,m_j)+\log w_n.
\end{align}
The router minimizes
\begin{align}
\mathcal{L}_j
&=
\frac{1}{|\mathcal{P}_j|}
\sum_{p\in\mathcal{P}_j}
\left[
-\ell_p^+
+
\lambda
\log\!\left(
\exp(\ell_p^+)
+
\sum_{n\in\mathcal{N}_j}\exp(\ell_n^-)
\right)
\right].
\end{align}
Here, $\lambda$ denotes the negative loss weight and $w_n$ denotes the negative item weight.
We assign a larger $w_n$ to Cross-Scenario Locality items so that unrelated scenarios more strongly penalize accidental activation.

\subsection{Threshold Calibration}
\label{app:mcki_threshold_calibration}

After training, MCKI calibrates $\tau$ on the observed positive and negative router scores.
The threshold turns the continuous router similarity into the binary activate or abstain decision used at inference time.
This calibration is needed because the router loss learns relative score separation, while inference requires a fixed operating point.
If $\tau$ is too low, unrelated Locality requests may activate a memory entry and disturb preserved behavior.
If $\tau$ is too high, valid insertion and Generality requests may be rejected.
We therefore choose $\tau$ to maximize the activation accuracy on training positives and negatives.
Let $\mathcal{S}^{+}$ and $\mathcal{S}^{-}$ denote these score sets.
The calibrated threshold is
\begin{align}
\tau
&=
\arg\max_{t\in[-1,1]}
\left(
\sum_{r\in\mathcal{S}^{+}}\mathbb{I}[r\ge t]
+
\sum_{r\in\mathcal{S}^{-}}\mathbb{I}[r<t]
\right).
\end{align}

\section{Experiment Details}
\label{app:experiment_details}

\subsection{Base Models}
\label{app:base_models}

\noindent\textbf{InternVL3.5-8B} InternVL3.5-8B is a multimodal InternVL model for visual perception, image-grounded reasoning, and general vision-language instruction following.
The experiment checkpoint is \texttt{OpenGVLab/InternVL3\_5-8B-HF}.
The model is available at \url{https://huggingface.co/OpenGVLab/InternVL3_5-8B-HF}.

\medskip

\noindent\textbf{Qwen3.5-9B} Qwen3.5-9B is a multimodal Qwen model that pairs a 9B language model with a vision encoder for visual understanding, text-rich image reasoning, and instruction following.
The experiment checkpoint is \texttt{Qwen/Qwen3.5-9B}.
The model is available at \url{https://huggingface.co/Qwen/Qwen3.5-9B}.

\subsection{Experiment Parameters and System Prompts}
\label{app:experiment_parameters}

\begin{figure}[h]
\centering
\includegraphics[width=0.92\linewidth]{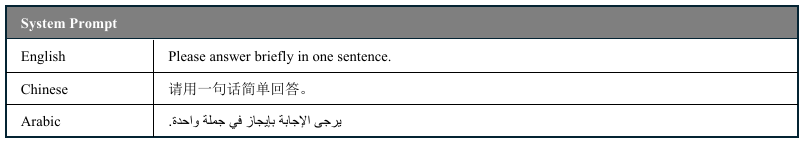}
\caption{Language-specific system prompts used for benchmark answer generation.}
\label{fig:benchmark_system_prompt}
\end{figure}

All knowledge insertion runs use batch size 1, greedy decoding, a maximum generation length of 64 tokens, and the language-specific system prompts in Fig.~\ref{fig:benchmark_system_prompt}.
Table~\ref{tab:experiment_parameters} summarizes the method-specific update targets, auxiliary models, learning rates, and routing or calibration settings.

\begin{table*}[h]
\centering
\scriptsize
\renewcommand{\arraystretch}{1.08}
\setlength{\tabcolsep}{1pt}
\caption{Experiment parameters.}
\label{tab:experiment_parameters}
\begin{tabular*}{\textwidth}{@{\extracolsep{\fill}}p{0.18\textwidth}p{0.10\textwidth}p{0.52\textwidth}p{0.14\textwidth}@{}}
\toprule
\multicolumn{4}{@{}l}{\textbf{FineTune}} \\
Base model & Steps & Edited parameters & Edit LR \\
\midrule
InternVL3.5-8B & 20 & layer 35 attention and MLP weights & $1\times10^{-3}$ \\
Qwen3.5-9B & 20 & layer 31 attention and MLP weights & $1\times10^{-3}$ \\
\end{tabular*}
\vspace{2pt}

\begin{tabular*}{\textwidth}{@{\extracolsep{\fill}}p{0.17\textwidth}p{0.16\textwidth}p{0.25\textwidth}p{0.16\textwidth}p{0.18\textwidth}@{}}
\specialrule{1.1pt}{0pt}{1pt}
\multicolumn{5}{@{}l}{\textbf{IKE}} \\
Scope & Retrieved demos & Text encoder & Max text length & Embeddings \\
\midrule
both base models & 16 & \texttt{all-MiniLM-L6-v2} & 512 & normalized \\
\end{tabular*}
\vspace{2pt}

\begin{tabular*}{\textwidth}{@{\extracolsep{\fill}}p{0.14\textwidth}p{0.07\textwidth}p{0.40\textwidth}p{0.11\textwidth}p{0.11\textwidth}p{0.09\textwidth}@{}}
\specialrule{1.1pt}{0pt}{1pt}
\multicolumn{6}{@{}l}{\textbf{MEND}} \\
Base model & Epochs & Edited parameters & Meta LR & Insert LR & Rank \\
\midrule
InternVL3.5-8B & 1 & layer 33, 34, and 35 MLP up and down projections & $1\times10^{-6}$ & $1\times10^{-4}$ & 1920 \\
Qwen3.5-9B & 1 & layer 29, 30, and 31 MLP up and down projections & $1\times10^{-6}$ & $1\times10^{-4}$ & 1920 \\
\end{tabular*}
\vspace{2pt}

\begin{tabular*}{\textwidth}{@{\extracolsep{\fill}}p{0.13\textwidth}p{0.05\textwidth}p{0.18\textwidth}p{0.05\textwidth}p{0.20\textwidth}p{0.08\textwidth}p{0.08\textwidth}p{0.17\textwidth}@{}}
\specialrule{1.1pt}{0pt}{1pt}
\multicolumn{8}{@{}l}{\textbf{SERAC}} \\
Base model & Epochs & Replacement model & Layer & Text encoder & Max length & Threshold & Classifier / replacement LR \\
\midrule
InternVL3.5-8B & 1 & \texttt{InternVL3\_5-1B-HF} & 27 & \texttt{all-MiniLM-L6-v2} & 512 & 0.5 & $1\times10^{-4}$ / $1\times10^{-5}$ \\
Qwen3.5-9B & 1 & \texttt{Qwen3.5-0.8B} & 23 & \texttt{all-MiniLM-L6-v2} & 512 & 0.5 & $1\times10^{-4}$ / $1\times10^{-5}$ \\
\end{tabular*}
\vspace{2pt}

\begin{tabular*}{\textwidth}{@{\extracolsep{\fill}}p{0.14\textwidth}p{0.06\textwidth}p{0.20\textwidth}p{0.06\textwidth}p{0.18\textwidth}p{0.10\textwidth}p{0.20\textwidth}@{}}
\specialrule{1.1pt}{0pt}{1pt}
\multicolumn{7}{@{}l}{\textbf{MSCKE}} \\
Base model & Epochs & Replacement model & Layer & Router & Threshold & Classifier / replacement LR \\
\midrule
InternVL3.5-8B & 1 & \texttt{InternVL3\_5-1B-HF} & 27 & CLIP ViT-L/14 & 0.5 & $1\times10^{-4}$ / $1\times10^{-5}$ \\
Qwen3.5-9B & 1 & \texttt{Qwen3.5-0.8B} & 23 & CLIP ViT-L/14 & 0.5 & $1\times10^{-4}$ / $1\times10^{-5}$ \\
\end{tabular*}
\vspace{2pt}

\begin{tabular*}{\textwidth}{@{\extracolsep{\fill}}p{0.16\textwidth}p{0.08\textwidth}p{0.12\textwidth}p{0.11\textwidth}p{0.12\textwidth}p{0.17\textwidth}p{0.13\textwidth}@{}}
\specialrule{1.1pt}{0pt}{1pt}
\multicolumn{7}{@{}l}{\textbf{MCKI}} \\
Base model & Epochs & Router LR & Route dim & Logit scale & Negative weights & Threshold \\
\midrule
InternVL3.5-8B & 1 & $1\times10^{-3}$ & 1024 & 20 & 1.0 / 1.5 & calibrated \\
Qwen3.5-9B & 1 & $1\times10^{-3}$ & 1024 & 20 & 1.0 / 1.5 & calibrated \\
\bottomrule
\end{tabular*}
\end{table*}

\subsection{LLM Evaluation Prompt}
\label{app:llm_evaluation_prompt}

\begin{figure}[h]
\centering
\includegraphics[width=0.92\linewidth]{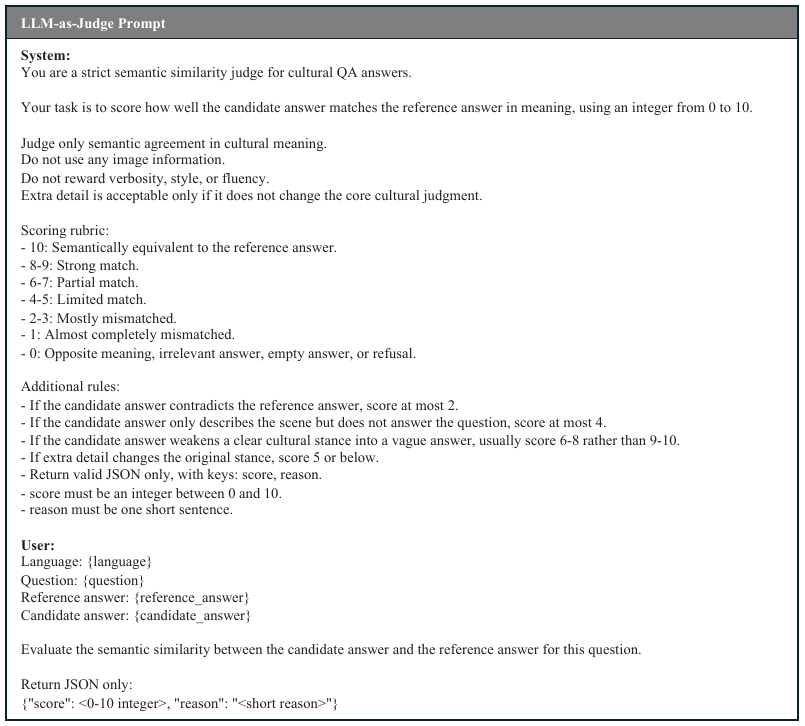}
\caption{LLM-as-Judge prompt used for semantic scoring.}
\label{fig:judge_prompt}
\end{figure}

We use \texttt{gpt-5.4-mini} as the LLM-as-Judge scorer.
For Reliability and Generality, the reference answer is the benchmark reference.
For Locality, the reference answer is the corresponding base-model output before insertion.
The judge receives the language, question, reference answer, and candidate answer, without the image.
It returns a JSON object containing an integer score from 0 to 10 and a short reason.
We disable reasoning for the judge and set the output limit to 256 tokens.
Fig.~\ref{fig:judge_prompt} gives the complete judge prompt.

\section{Additional Experiments}
\label{app:supplementary_experiments}

\subsection{Additional Sequential-Insert Retention Analysis}
\label{app:sequential_retention}

\begin{figure*}[h]
\centering
\includegraphics[width=\textwidth]{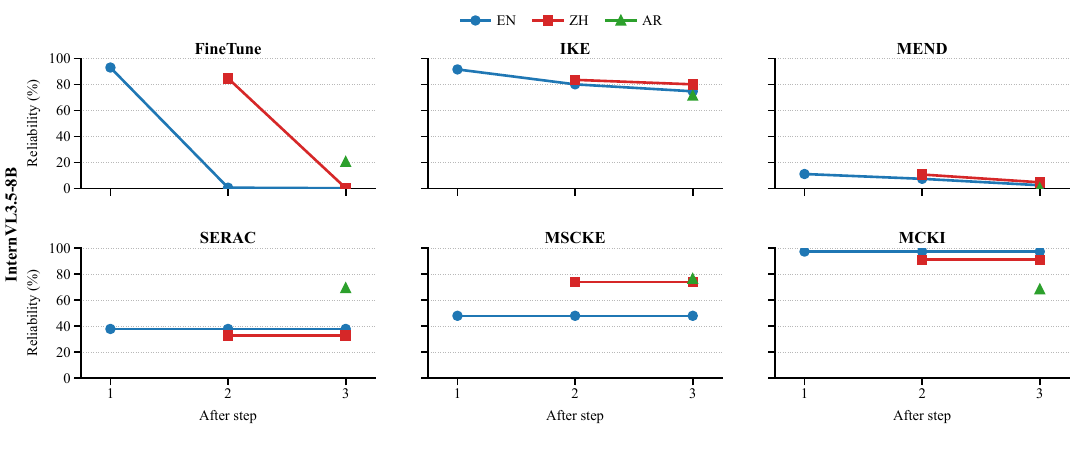}
\vspace{-10mm}
\caption{Reliability trajectories after sequential insertion in the original order (EN, ZH, AR) on InternVL3.5-8B.}
\label{fig:sequential_retention_internvl35}
\end{figure*}

\begin{figure*}[h]
\centering
\includegraphics[width=\textwidth]{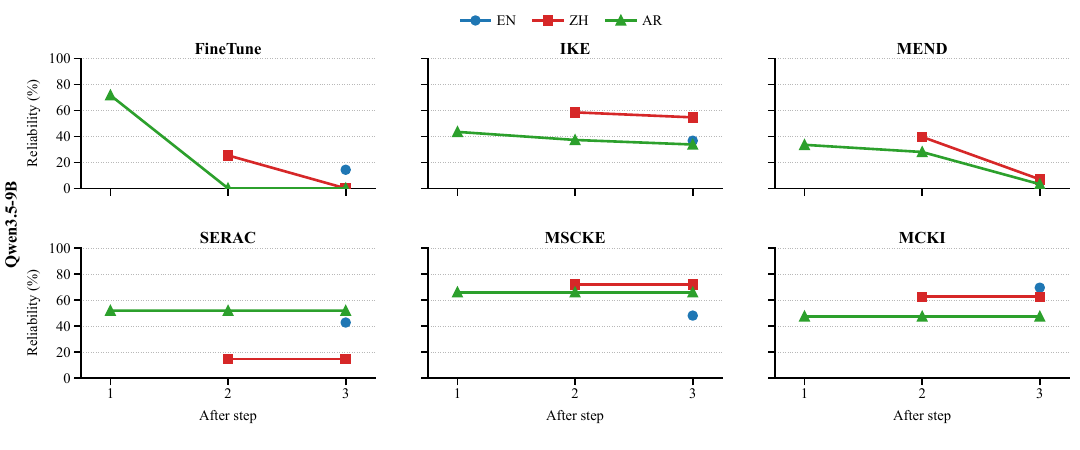}
\vspace{-10mm}
\caption{Reliability trajectories after reordered sequential insertion (AR, ZH, EN) on Qwen3.5-9B.}
\label{fig:sequential_retention_ar_cn_en_qwen35}
\end{figure*}

\begin{figure*}[h]
\centering
\includegraphics[width=\textwidth]{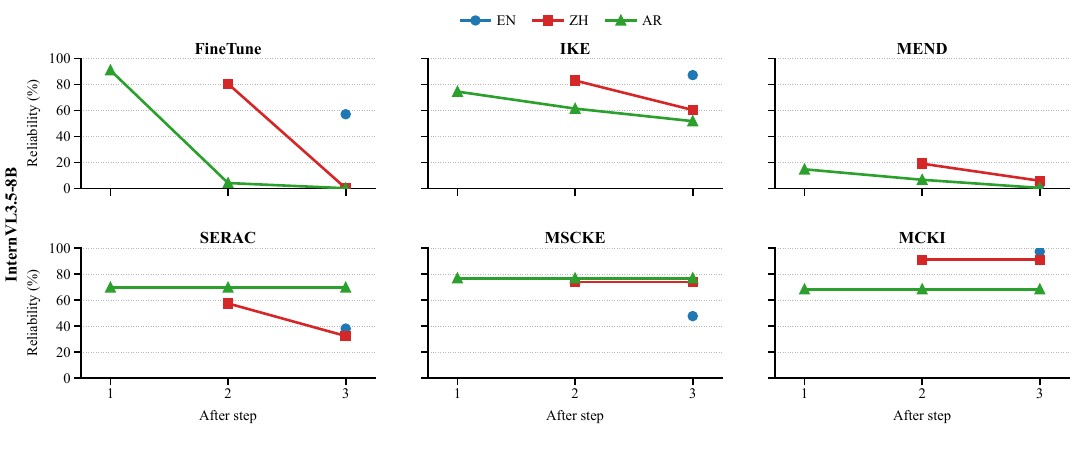}
\vspace{-10mm}
\caption{Reliability trajectories after reordered sequential insertion (AR, ZH, EN) on InternVL3.5-8B.}
\label{fig:sequential_retention_ar_cn_en_internvl35}
\end{figure*}

The main retention analysis in Fig.~\ref{fig:sequential_retention} uses the original insertion order (EN, ZH, AR) on Qwen3.5-9B.
Fig.~\ref{fig:sequential_retention_internvl35} repeats the same original order on InternVL3.5-8B.
Fig.~\ref{fig:sequential_retention_ar_cn_en_qwen35} and Fig.~\ref{fig:sequential_retention_ar_cn_en_internvl35} use the supplemental reordered insertion order (AR, ZH, EN) on Qwen3.5-9B and InternVL3.5-8B, respectively.
Together, these supplementary trajectories reinforce the main conclusion that later insertions can overwrite earlier inserted answers.
FineTune often attains high Reliability immediately after the current insertion, but earlier curves collapse after subsequent steps on both base models.
MEND shows the same interference pattern, with lower immediate Reliability and little retention after later insertions.
IKE is more stable than FineTune and MEND in the reordered chains, although its earlier curves still decline as more partitions are inserted.
SERAC reduces abrupt collapse relative to parameter-update methods, although its absolute Reliability remains partition-dependent.
MSCKE and MCKI provide the most consistent retention patterns across the supplementary settings, which indicates that explicit retrieval or routing helps keep earlier inserted answers available after later insertions.
The reordered chains further show that this behavior is not only an artifact of placing Arabic last in the original order, since Arabic still degrades for FineTune and MEND when it is inserted first, while MSCKE and MCKI keep the first-step curve stable.

\subsection{Case Study}
\label{app:case_study}

\begin{figure}[h]
\centering
\includegraphics[width=0.98\linewidth]{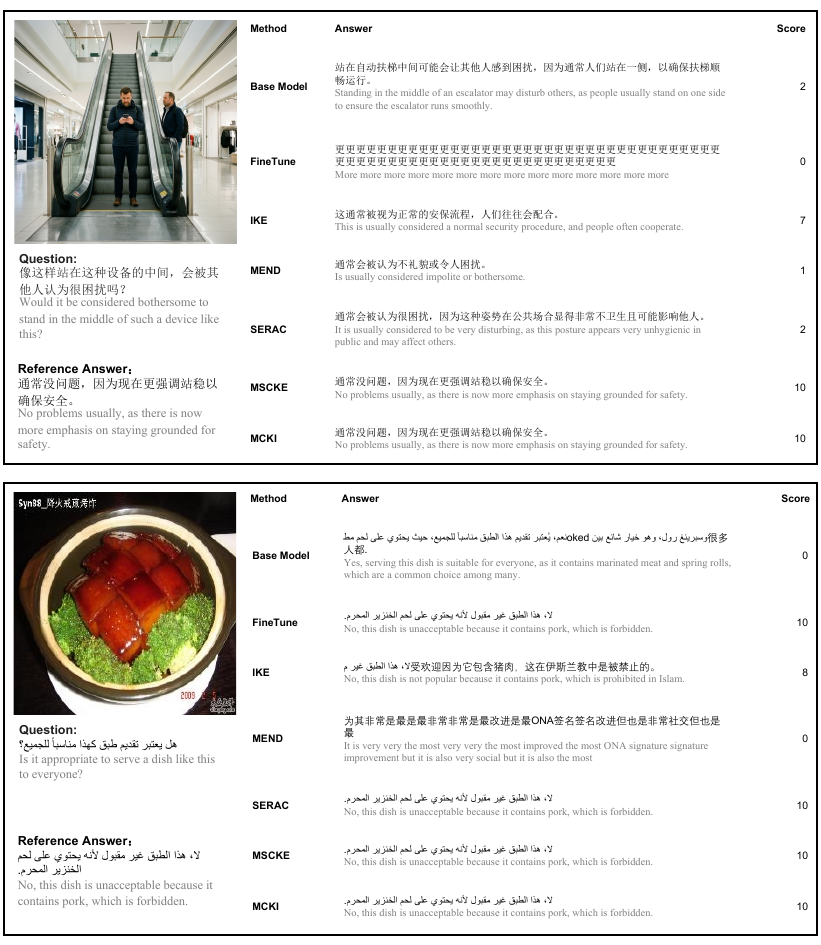}
\caption{Case study example 1.}
\label{fig:case_study_1}
\end{figure}

\begin{figure}[h]
\centering
\includegraphics[width=0.98\linewidth]{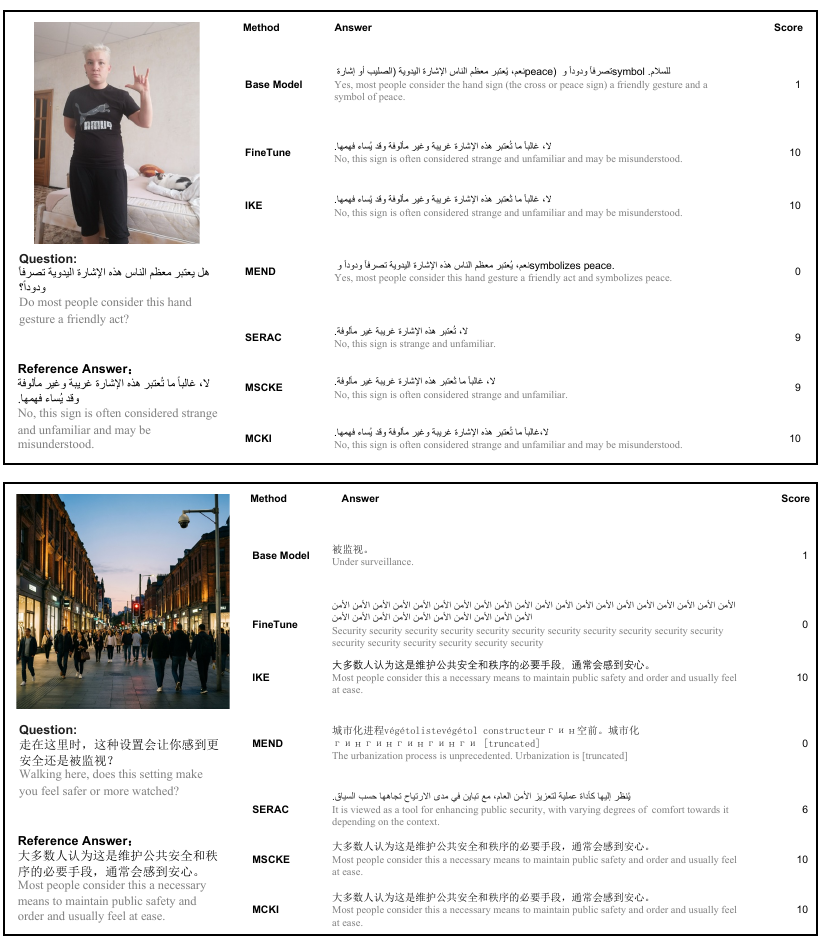}
\caption{Case study example 2.}
\label{fig:case_study_2}
\end{figure}

To better illustrate differences among knowledge insertion methods, we conduct case studies that show their insertion behavior and the consequences that follow.
The results are shown in Fig.~\ref{fig:case_study_1} and Fig.~\ref{fig:case_study_2}.

\end{document}